%% file: main.tex
\documentclass[conference,a4paper]{IEEEtran}
\IEEEoverridecommandlockouts

\usepackage{silence}
\usepackage{cite}

\usepackage{amsmath,amssymb,amsthm,fixmath}

\usepackage{color,colortbl}
\usepackage{xcolor}

\usepackage[inline]{enumitem}

\usepackage{siunitx}
\DeclareSIUnit{\dBm}{dBm}

\usepackage{graphicx}
\usepackage[labelformat=simple]{subcaption}

\usepackage{epstopdf}
\usepackage{lipsum}

\usepackage{cuted}
\usepackage{multirow,booktabs}
\usepackage{diagbox}
\usepackage{makecell}
\usepackage{hhline}
\usepackage{array} 
\newcolumntype{x}{!{\vrule width 2px}}
\newcolumntype{y}{!{\vrule width 1.5px}}

\usepackage{optidef}

\usepackage{nohyperref} 

\usepackage[shortcuts,acronym,automake]{glossaries}
\input{glossary}

\usepackage{tcolorbox}
\tcbuselibrary{many}
\newtcbtheorem[]{alg}{Algorithm}{fonttitle=\bfseries}{alg}
\usepackage[norelsize,linesnumbered,ruled]{algorithm2e}
\SetKwRepeat{Do}{do}{while}
\makeatletter
\newcommand{\removelatexerror} {\let\@latex@error\@gobble}
\makeatother

\newcommand{\subscript}[1]{_{\mathrm{#1}}}
\newcommand{\diff}{\text{d}}

\usepackage[normalem]{ulem}

\usepackage[textsize=tiny,colorinlistoftodos]{todonotes}
\makeatletter
\define@key{todonotes}{bh}[]{% Comment by 
	\setkeys{todonotes}{author=\textbf{Bin}, color=lime!30}}%
\makeatother

\tikzstyle{note}=[rectangle, minimum width=3cm, draw = none, fill = none, minimum width = 1.5cm, anchor=center, align=left]
\tikzstyle{block}=[rectangle, draw, line width=1pt, fill = none, minimum width = 1cm, minimum height = 0.75cm, anchor=center, inner sep = 0.5mm, align=center]
\tikzstyle{arrow} = [thick,->,>=stealth]

\newif\ifreviewmode
\reviewmodetrue % Set to true to enable review mode
\ifreviewmode
\else
  \renewcommand{\todo}[1]{} % hide todo notes
\fi

\begin{document}
\let\url\gobble

\title{Quotation-Based Data Retention Mechanism for Data Privacy in LLM-Empowered Network Services}
% \title{Buyer-Initiated Auction Mechanism for\\Data Redemption in Machine Unlearning}
% \title{Buyer-Initiated Auction for Data Redemption}
\author{
	\IEEEauthorblockN{
		Bin~Han\IEEEauthorrefmark{1},
        Di~Feng\IEEEauthorrefmark{2},
		Zexin~Fang\IEEEauthorrefmark{1},
		Jie~Wang\IEEEauthorrefmark{3},
        and~Hans~D.~Schotten\IEEEauthorrefmark{1}\IEEEauthorrefmark{4}
	}
	\IEEEauthorblockA{
		% \IEEEauthorrefmark{1}Division of Wireless Communications and Radio Positioning, RPTU University Kaiserslautern-Landau, Kaiserslautern, Germany\\
        % \IEEEauthorrefmark{2}Finance Department, Dongbei University of Finance and Economics, Dalian, China\\ 
		% \IEEEauthorrefmark{3}Department of Electical and Computer Engineering, Tongji University, Shanghai, China\\
        % \IEEEauthorrefmark{4}Intelligent Networks, German Research Center for Artificial Intelligence (DFKI), Kaiserslautern, Germany
		\IEEEauthorrefmark{1}RPTU University Kaiserslautern-Landau, Kaiserslautern, Germany\\
        \IEEEauthorrefmark{2}Dongbei University of Finance and Economics, Dalian, China\\ 
		\IEEEauthorrefmark{3}Tongji University, Shanghai, China\\
        \IEEEauthorrefmark{4}German Research Center for Artificial Intelligence (DFKI), Kaiserslautern, Germany
	}
}

\maketitle

\begin{abstract}
The deployment of \acp{llm} for next-generation network optimization introduces novel data governance challenges. \acp{mno} increasingly leverage generative \ac{ai} for traffic prediction, anomaly detection, and service personalization, requiring access to users' sensitive network usage data--including mobility patterns, traffic types, and location histories. Under the \ac{gdpr}, the \ac{ccpa}, and similar regulations, users retain the right to withdraw consent and demand data deletion. However, extensive machine unlearning degrades model accuracy and incurs substantial computational costs, ultimately harming network performance for all users. We propose an iterative price discovery mechanism enabling \acp{mno} to compensate users for data retention through sequential price quotations. 
The server progressively raises the unit price for retaining data while users independently determine their supply at each quoted price. This approach requires no prior knowledge of users' privacy preferences and efficiently maximizes social welfare across the network ecosystem.
% The rapid growth of \ac{ai} has raised privacy concerns over user data, leading to regulations like the \ac{gdpr} and the \ac{ccpa}. With the essential toolbox provided by machine unlearning, \ac{ai} service providers are now able to remove user data from their trained models as well as the training datasets, so as to comply with such regulations. However, extensive data redemption can be costly and degrade model accuracy. To balance the cost of unlearning and the privacy protection, we propose a buyer-initiated auction mechanism for data redemption, enabling the service provider to purchase data from willing users with appropriate compensation. This approach does not require the server to have any a priori knowledge about the users' privacy preference, and provides an efficient solution for maximizing the social welfare in the investigated problem.
\end{abstract}


\glsresetall

\section{Introduction}\label{sec:intro}
The integration of \acp{llm} and generative \ac{ai} into next-generation mobile networks has emerged as a key enabler for intelligent network management and operations~\cite{BSA+2026survey}. 
\acp{mno} deploy \acp{llm} for predictive maintenance, automated troubleshooting, traffic forecasting, and personalized service 
delivery. These applications require training on extensive user data--including cell handover records, packet inspection logs, geographic mobility traces, and application usage patterns--to achieve acceptable accuracy.
% The rapid development of \ac{ai} and \ac{ml} technologies over the past decade has brought about a prospery of intelligent applications and a new era of digital economy, recently supplemented by the explosive blowout of \acp{llm}. As the feul of this new techno-economic paradigm, high-quality data has become a valuable asset and in thirst demand. In practice, the data used for training \ac{ml} models is often collected from the users of the applications. 

However, the data collection process may raise privacy concerns, as the data may contain sensitive information about the users, and the models therewith trained may leak or abuse such information against the users' will. In network operations contexts, this tension intensifies: user data fed into \ac{llm} training directly influences \ac{qos} for the entire user base, creating 
interdependencies between individual privacy and collective network performance.

Being aware of these privacy concerns, data regulations have been established in various countries and regions, represented by the \ac{gdpr} in the European Union~\cite{gdpr}, the \ac{ccpa} in California~\cite{ccpa}, and the \ac{pipl} in China~\cite{pipl}. Such regulations generally require the service providers to obtain the users' consent before collecting their data, and to provide the users with the right to delete their data afterwards.

To enable removing user data from \ac{llm} training pipelines--both from stored datasets and deployed models--\acp{mno} must implement machine unlearning~\cite{XZZ+2023machine}. Despite technical feasibility, excessive unlearning creates dual challenges 
for network operators: \begin{enumerate*}[label=(\arabic*)]
	\item computational overhead in retraining resource-
intensive \acp{llm}, and
	\item degraded predictive accuracy that cascades into worse 
traffic management, increased latency, and higher outage rates~\cite{CC2024price}.
\end{enumerate*}
A rigid unlearning policy paradoxically harms both \ac{mno} operational efficiency and user experience, necessitating market-based alternatives.
% To enable removing user data from the server in both senses of the training dataset and the trained model, the technologies of \emph{machine unlearning} have been developed and attracted much attention~\cite{XZZ+2023machine}. Despite the technical success, however, an excessive unlearning of user data can be not only computationally costly for the \ac{ai} service provider, but also significantly degrading the model's accuracy, which also harm the users' quality of experience~\cite{NOD+2022markov}. Therefore, a rigid unlearning policy may be the long run lead against both interests of the service provider and the users.

To address this issue, a new data redemption framework in on call for, which shall balance the cost of unlearning and the privacy protection. One of the appealing approaches is data monetization, i.e., the \ac{ai} service provider encourages its users not to redeem their data by offering them a compensation. This allows the users to flexibly choose between reserving and selling their data upon their own will and interest, while the service provider benefits from the reduced cost of unlearning and better model accuracy.

The crucial challenge in this framework is to determine the price of data. Although data pricing has been extensively studied in the context of data marketplaces~\cite{ZBL2023survey}, existing works generally deal with the scenario of active data trading for model training, focusing on the quality of data and the model performance, while the data redemption and unlearning scenario is barely explored. This substential difference leads to an untolarable inapplicability of the existing pricing models in the data redemption context. For example, some works such like~\cite{ZYZ2024price} simply ignores the privacy loss of users when they sell data to the server, while other works such like~\cite{NZW+2020online}, though taking privacy into account, generally do not capture the computational cost caused by unlearning.

A piorneering work on the topic of pricing in data redemption is presented in~\cite{CC2024price}, where the authors propose a novel incentive mechanism for machine unlearning. Their model considers the server's cost of unlearning regarding both computation and accuracy degradation, and also takes into account the users' privacy utility. A two-stage optimization problem is formulated, where on the first stage the server determines an optimal unit price for user data, and on the second stage every user determin its own optimal amount of data to sell.

Despite the novelty and inspiration of the work in~\cite{CC2024price}, there are still some limitations in it. First, the server's optimal pricing relies on certain assumptions in the distribution of users' privacy parameters, which may not hold in practical scenarios. Second, the two-stage pricing mechanism always returns one uniform price for every data batch for all users. Considering the heterogeneity of users' privacy concerns and the non-linearity of user's privacy utility regarding the redeemed data amount, this is possibly leading to a suboptimum.

In this work, we propose an iterative price discovery mechanism based on 
sequential quotations to address the above limitations. The server (\ac{mno}) 
progressively quotes higher unit prices for data retention, and users 
independently determine their willingness to supply data at each price point. 
This quotation-based approach enables a flexible and progressive pricing to convey the non-linear cost/utility functions of both the server and the users, and requires no prior knowledge on the server side about the distribution of users' privacy parameters.
% \DF{It is better to refer to it as an ascending price auction rather than an English auction. Typically, the term English auction implies that each bidder knows other bidders' histories.}
% In a multi-round auction, the server gradually raises the unit price for purchasing data, and each user independently decides how much data it is willing to sell at the offered price. After the users claiming, the server determines how much data it is willing to buy at the current price, before announcing the new price for the next round or closing the auction. This auction mechanism enables a flexible and progressive pricing to convey the non-linear cost/utility functions of both the server and the users, and requires no prior knowledge on the server side about the distribution of users' privacy parameters.

The remainding part of this paper is organized as follows. In Section~\ref{sec:model}, we present the system model, including the cost and utility functions of the server and the users, respectively. In Section~\ref{sec:analyses}, we analyze the server's incentive to purchase data and the users' incentive to sell data in an arbitrary quotation round. In Section~\ref{sec:approach}, we propose the iterative quotation-based price discovery mechanism, and discuss the optimal decision strategies for the server and the users. Subsequently, we numerically evaluate our proposed approach with simulation campaigns in Sec.~\ref{sec:evaluation}, before finally concluding the paper in Section~\ref{sec:conclusion}.

\section{System Model}\label{sec:model}
We model the interaction between an \ac{mno} deploying \acp{llm} for network optimization and its user base. The \ac{mno} (henceforth "server") operates an \ac{llm} trained on aggregated user data to perform tasks such as:
\begin{itemize}
	\item Traffic prediction: Forecasting bandwidth demand for proactive resource allocation;
	\item Anomaly detection: Identifying network faults or security threats;
	\item Service personalization: Recommending optimal connectivity plans.
\end{itemize}

Each user $i\in\mathcal{I}$ has contributed data di (measured in normalized units, 
e.g., number of mobility records or session logs). Under GDPR Article 17, 
users may request data deletion, triggering machine unlearning obligations. 
The following cost and utility models capture the economic tradeoffs.

\subsection{Server Cost Model}
Considering the cost model in \cite{CC2024price} that $C(x)=\alpha A(x)+ \beta T(x)$, where $x=\sum\limits_{i=1}^I x_i$, $x_i$ is the amount of data redemption for user $i\in\mathcal{I}$, $A(x)$ is the accuracy degradation of the model caused by unlearning, and $T(x)$ the computing time for executing the unlearning task. Note that $\alpha,A,\beta,T$ are all non-negative. Defining the total data amount for each user $i$ as $d_i$, and their sum $d=\sum\limits_{i=1}^Id_i$, according to \cite{CC2024price}:
\begin{align}
	A(x)&=A_1a^{A_2x}-A_3,\label{eq:A(x)}\\
	T(x)&=\begin{cases}
		0 & \text{if } x=0,\\
		T_0(d-x) & \text{if } x\in(0,d].
	\end{cases}\label{eq:T(x)}
\end{align}

From the server's perspective it is more convenient to focus on the amount of data to keep, i.e. to buy from the users, rather than the amount of data to redeem. So we define that $y_i=d_i-x_i$ and $y=d-x$, and rewrite the cost function as
\begin{align}
	C(y)&=\alpha A(y)+\beta T(y),\\
	A(y)&=A_1e^{A_2(d-y)}-A_3,\\
	T(y)&=\begin{cases}
		T_0y & \text{if } y\in[0,d),\\
		0 & \text{if } y=d.
	\end{cases}
\end{align}

\subsection{User Privacy Model}
Following \cite{CC2024price} we consider that for each user $i\in\mathcal{I}$ the utility of privacy is captured by
\begin{equation}
	U_i(x_i)=\lambda_i\ln(x_i+1),
\end{equation} 
where $\lambda_i$ is its privacy parameter. Similarly, focusing on the amount of data to keep on the server, we rewrite it As
\begin{equation}
	U_i(y_i)=\lambda_i\ln(d_i-y_i+1),
\end{equation}
which is concave and monotonically decreasing w.r.t. $y_i$.

\section{Incentive Analyses}\label{sec:analyses}
\subsection{Server's Incentive to Purchase Data}

\begin{figure}
	\centering
	\includegraphics[width=\linewidth]{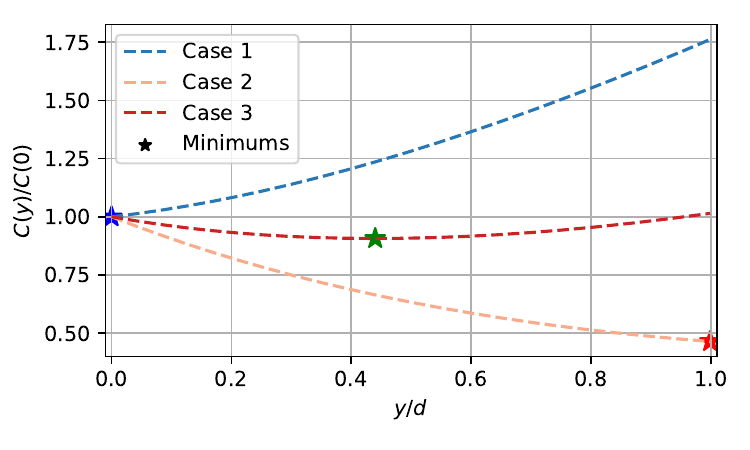}
	\caption{Different cases of the server's cost function.}
	\label{fig:server_cost_cases}
\end{figure}

The server has an incentive to keep as much data as to minimize the cost of unlearning. We denote this target amount of data to keep as $y\subscript{max}$. Note that $C(y)$ has one and only one jump discontinuity at $y=d$, for simplification of discussion we first focus on the continuous interval $y\in[0,d)$, and leave the case of $y=d$ for later discussion in Sec.~\ref{subsec:discontinuity}. Thus, $C(y)$ is in general a convex function over $y\in[0,d)$, while its monoticity, depending on the specific parameter settings, may have three diffferent cases, as shown in Fig.~\ref{fig:server_cost_cases}:
\begin{enumerate}[label=\arabic*):]
	\item monotonically increasing,
	\item monotonically decreasing, or
	\item non-monotone.
\end{enumerate}
In the first case, trivially, the server will not keep any data and $y\subscript{max}=0$. In the second case, the server tend to keep all data and $y\subscript{max}\to d^-$. In the third case, $y\subscript{max}$ can be obtained by looking for the zero derivative:
\begin{align}
	\left.\frac{\diff C(y)}{\diff y}\right\vert_{y\subscript{max}}&=(-\alpha A_1A_2\ln a)a^{A_2(d-y\subscript{max})}+\beta T_0=0,\\
	\Rightarrow y\subscript{max}&=d+\frac{1}{A_2}\log_a\left(\frac{\alpha A_1A_2\ln a}{\beta T_0}\right).\label{eq:ymax}
\end{align}

Now assume that the server has already managed to keep $y<y\subscript{max}$ amount of data, and offers to buy more data from users at a price $B$ per unit amount of data. The server's \emph{demand} $\eta$ will be a function of the condition $(y,B)$, which is:
\begin{align}
		&\eta(y,B)=\underset{\delta\in[0,y\subscript{max}-y]}{\arg\max}\left[C(y)-C(y+\delta)-B\delta\right]\label{eq:demand}\\
		=&\underset{\delta\in[0,y\subscript{max}-y]}{\arg\max}\left[\alpha A_1e^{A_2(d-y)}(1-e^{-A_2\delta)})-(\beta T_0+B)\delta\right].\nonumber
\end{align}

Especially, noticing that $C(y)$ is always monotonic over the interval $y\in[0,y\subscript{max})$, there is a unique maximum price $B\subscript{all}(y)$ to support a ``buy all'' strategy:
\begin{equation}
	\begin{split}
		&B\subscript{all}(y)=\frac{C(y\subscript{max})-C(y)}{y\subscript{max}-y}\\
		% =&\frac{\alpha A_1e^{A_2(d-y\subscript{max})}-\alpha A_1e^{A_2(d-y)}}{y\subscript{max}-y}+\beta T_0\\
		=&\frac{\alpha A_1e^{A_2d}\left(e^{-A_2y\subscript{max}}-e^{-A_2y}\right)}{y\subscript{max}-y}+\beta T_0.
	\end{split}
\end{equation}
With $y$ data already bought from the users, given any price $B\leqslant B\subscript{all}(y)$, the server will buy as much data from the users as possible, until it reaches the maximum amount $y\subscript{max}$ of data to keep. Note that in case of $y$-increasing cost $C$ (Case 1 in Fig.~\ref{fig:server_cost_cases}), $B\subscript{all}\leqslant 0$ and $y\subscript{max}=0$.

\subsection{User's Incentive to Sell Data}
Given the price $B$ offered by the server, each user $i$ can decide how much data it will sell to the server, based on how much data $y_i$ it has already sold previously. The change of privacy utility of selling $\delta_i$ more data is
\begin{equation}
	\begin{split}
		&\Delta U_i(y_i,\delta_i)=U_i(y_i+\delta_i) - U_i(y_i)\\
		=&\lambda_i\ln\left(1-\frac{\delta_i}{d_i-y_i+1}\right),
	\end{split}
\end{equation}
so the minimum price to convince user $i$ to sell $\delta_i$ more data must compensate this utility loss:
\begin{equation}
	\begin{split}
		&B_{\text{min},i}(y_i,\delta_i)=\frac{-\Delta U_i(y_i,\delta_i)}{\delta_i}\\
		=&\frac{\lambda_i}{\delta_i}\ln\left(1+\frac{\delta_i}{d_i-y_i+1-\delta_i}\right).
	\end{split}
\end{equation}
It is trivial that for all $y_i\in[0,d_i)$ and $\delta_i\geqslant 0$:
\begin{align}
	\frac{\partial}{\partial y_i}B_{\text{min},i}(y_i,\delta_i)&>0,\\
	\frac{\partial}{\partial \delta_i}B_{\text{min},i}(y_i,\delta_i)&>0,
\end{align}
so there is a minimal price for user $i$ to sell any amount of data in addition to the already-sold $y_i$:
\begin{equation}
	\lim\limits_{\delta_i\to0}B_{\text{min},i}(y_i,\delta_i)=\frac{\lambda_i}{d_i-y_i+1},
\end{equation}
and a minimal price for user $i$ to sell all the remaining data:
\begin{equation}
	B_{\text{all},i}(y_i)=B_{\text{min},i}(y_i,d_i-y_i)=\frac{\lambda_i}{d_i-y_i}\ln(d_i-y_i+1).  
\end{equation}

Moreover, given that user $i$ has already sold $y_i$ in priori, there is a maximal amount of data $q_i$ it is willing to sell to the server at price $p$, which we call $i$'s \emph{supply}:
\begin{equation}
	\begin{split}
		&q_i(y_i,B)=\underset{\delta\in[0,d_i-y_i]}{\arg\max}\left[p\delta+\Delta U_i(y_i,\delta)\right].\\
		=&\underset{\delta\in[0,d_i-y_i]}{\arg\max}\left[p\delta+\lambda_i\ln\left(1-\frac{\delta}{d_i-y_i+1}\right)\right].
	\end{split}   \label{equ:optimalselling}
\end{equation}

\section{Quotation-Based Price Discovery Mechanism}\label{sec:approach}
% \section{Multi-Round Auction Mechanism}\label{sec:approach}
\subsection{Quotation-Based Price Discovery Protocol}
To balance the server's cost of unlearning and the users' privacy utility, 
we propose an iterative price discovery mechanism through sequential 
quotations. The protocol operates as follows:
% \subsection{Buyer-Initiated English Auction}
% To balance the server's cost of unlearning and the users' privacy utility, we propose a buyer-initiated ascending auction mechanism, which can be briefly summarized as follows:
\begin{enumerate}
	\item  At the beginning of each round $t$, the server updates its price $B^t$ for buying unit data from the users. We focus in this study on the ascending quotation format, where the data price is increasing w.r.t. time, i.e., for all $t<\tau$, $B^t<B^\tau$. It also estimates its demand for data $\eta^t=\eta(y^t, B^t)$, which is \emph{unknown} to the users. If the demand drops to zero, the quotation is terminated. Otherwise, the price is quoted to all users.\label{step:quote_price}
	\item Each user then independently decides how much data it can supply at this price.
	\item The server purchases the data from all users who are willing to sell at the quoted price. If the total supply exceeds its demand, it will only purchase its demanded amount of data. Otherwise, it will buy all data supplied.
	\item If the server has collected the target amount of data $y\subscript{max}$, the quotation is terminated. Otherwise, $t\gets t+1$, and go back to step \ref{step:quote_price}) for the next round.
\end{enumerate}

For the sake of low communication overhead and privacy protection, our approach is to let the users make decisions independently from each other. Thus, each user $i$ is considered \emph{blind} in the sense that at each period $t$, $i$ only knows the current data price $B^t$, its own privacy parameter $\lambda_i$, its own data amount $d_i$, and its own history of data selling $(y_i^\tau)_{\tau<t}$. As we are considering an ascending quotation, the price quoted by the server along the time assembles an ascending series $(B^\tau)_{1\leqslant\tau<\infty}$. 

\subsection{Users' Selling Strategy}
Noticing the $t$-monotonically increasing price $B^t$, when a user has incentive to sell data in round $t$, it has also the alternative option of keeping the data to sell it at higher price in a future round. Herewith we discuss the optimal selling strategy of users in such multi-round quotation.

Suppose the quotation is now at the beginning of round $t$. Each user $i$'s current remaining data that can be sold is $x_i^t\geqslant 0$. Given the information that $i$ knows, $i$ forms a belief $\mu_i^t\in [0,1]$, which represents a \emph{subjective} probability that the quotation is terminated after round $t$. Thus, $\mu_i^t$ is a function about server's remaining demand $\eta^t$ at $t$, and the supply $q_j^t$ of all users $j\in\mathcal{I}$. 

Moreover, at $t$, since $i$ only knows its own history $(y_i^\tau)_{\tau<t}$, and that $y_j^\tau$ is non-decreasing about $\tau$ for all $j\in\mathcal{I}$, $\mu_i^t$ is increasing with $i$'s own sold data $y_i^t$, and the collection $(\mu_i^\tau)_{1\leqslant \tau<\infty}$ is (point-wisely) decreasing with $\tau$, for any $(y_i^\tau)_{1\leqslant \tau<\infty}$.

%Given $B^t$, if there is no next period, then $i$ should sell as much as he can (see (\ref{equ:optimalselling})).However, if $i$ sells a mound of data $\hat{y}_i^t$ and there is a next period, with higher price $\beta_i^{t+1}>B^t$, then $i$ faces a regret as he should sell the amount of data $\hat{y}_i^t$ with $\beta_i^{t+1}$, where $\beta_i^{t+1}$ is ($i$'s subjective) expected price at $t+1$, given (1) $i$'s supply at $t$ being zero, and (2) his belief at $t$, $\mu_i^t$. Thus, $i$ only sells data at $t$ if and only if the expected payoff of waiting for the next period is lower, i.e.,
%\begin{equation}
 %   \mu_i^t(Q_i^t)\cdot V(Q_i^t,B^{t})\geqslant [1-\mu_i^t(0)] \cdot E_{t+1}[V(Q_i^{t+1},\beta_i^{t+1})] \label{equ:cutoff}
%\end{equation}
 % and $Q_i^{t+1}$ is his optimal level of selling at price $\beta_i^{t+1}$.

Let $t_i$ be the first period that $i$ decides to sell, where according to the data price $B^{t_i}$, the optimal amount of selling for $i$ is 
automatically given by (\ref{equ:optimalselling}). Denote this amount of data be $q_i^{t_i}=q_i(0,B^{t_i})>0$. If $q_i^{t_i}=d_i$, it is straightfoward. Otherwise, supposing $q_i^{t_i}<d_i$, by the section of $t_i$ we have
\begin{equation}
	\mu_i^{t_i}(q_i^{t_i})\cdot V(q_i^{t_i},B^{t_i})\geqslant [1-\mu_i^{t_i}(0)] \cdot V(q_i^{t_i+1},B^{t_i+1}),\label{eq:better_sell_than_keep}
\end{equation} 
%\begin{equation}
 %   \mu_i^{t_i}(q_i^{t_i})\cdot V(q_i^{t_i},B^{t_i})\geqslant [1-\mu_i^{t_i}(0)] \cdot V(q_i^{t_i+1},B^{t_i+1}) \label{equ:cutoff}
%\end{equation}
where $V(y,B)$ is $i$'s benefit by selling the amount of data $y$ with price $B$, according to his utility. Thus, Eq.~\eqref{eq:better_sell_than_keep} implies that the expected payoff of selling at $t_i$ is higher than the the expected payoff of keeping data at $t_i$. In other words, as long as the users are blind about the server's demand and other users' trade record, they always perform the greedy strategy in selling data.

Furthermore, rearranging Eq.~\eqref{eq:better_sell_than_keep}, we have
\begin{equation}
	\frac{  \mu_i^{t_i} (q_i^{t_i})  }{[1-\mu_i^{t_i}(0)] }\geqslant \frac{V(q_i^{t_i+1},B^{t_i+1})}{V(q_i^{t_i},B^{t_i})}.	
\end{equation}
With the aforementioned monotonicity of $\mu_i$, we also have
\begin{equation}
	\frac{  \mu_i^{t_i+1} (q_i^{t_i+1})  }{[1-\mu_i^{t_i+1}(0)] }\geqslant \frac{  \mu_i^{t_i} (q_i^{t_i})  }{[1-\mu_i^{t_i}(0)] }.
\end{equation}
By the definitions of $q$ and $V$, $i$'s marginal benefit is decreasing, and hence
\begin{equation}
	\frac{V(q_i^{t_i+1},B^{t_i+1})}{V(q_i^{t_i},B^{t_i})}\geqslant \frac{V(q_i^{t_i+2},B^{t_i+2})}{V(q_i^{t_i+1},B^{t_i+1})}.	
\end{equation}
Thus, we can obtain that
\begin{equation}
	\frac{  \mu_i^{t_i} (q_i^{t_i})  }{[1-\mu_i^{t_i}(0)] }\geqslant  \frac{V(q_i^{t_i+2},B^{t_i+2})}{V(q_i^{t_i+1},B^{t_i+1})}.\label{eq:once_sell_never_back}
\end{equation}
Eq.~\eqref{eq:once_sell_never_back} implies, if $i$ starts to sell data at $t_i$, it will also choose to sell data in all future rounds, until having all its data sold. 
% As shown above, the reason here is simple: $i$'s belief and his marginal benefit are both decreasing.

% Therefore, given any increasing $(B^\tau)_{1\leqslant \tau<\infty}$, each user $i$'s optimal strategy can be identified by a pair: $t_i$ and $(q_i^\tau)_{t_1\leqslant \tau<\infty}$.

% \subsection{Optimal pricing}
% Here we analyze the server's pricing strategy. Recall that for any price list $B\equiv (B^\tau)_{1\leqslant \tau<\infty}$, each $i$'s optimal strategy is given by $(t_i(B^{t_i}), (q_i^\tau(B^\tau))_{t_1\leqslant \tau<\infty})$. It is without loss of generality to assume that for each $i$ and each $\tau<t_i$, $q_i^\tau=0$.

% Therefore, given any $B$, the server's (total amount of) data to keep is given by $y(B)\equiv d- q$, where $q=\sum_{i} \sum_{\tau}q_i^\tau$. Since $C$ is convex, one can find the optimal pricing (probably multi-valued) $B^*$ such that the server's cost is minimized.

\subsection{Handling the Oversupply}\label{subsec:oversupply}
Due to the wide-sense $B$-monotonicity of $\eta$ (decreasing) and $q_i$ (increasing) for all $i\in\mathcal{I}$, when the quoted price $B^t$ is sufficiently high, it may occur the case of $\eta^t<\sum\limits_{i=1}^Iq_i^t$, i.e. the total supply from users exceeds the server's demand. In this case, the server shall allocate its demand $\eta$ to the different selling users.

While neither the unit price $B^t$ nor the demand $\eta^t$ is influenced by the allocation, the server's purchase is independent therefrom, either. However, since the different users are of different characteristics $(y_i, \lambda_i)$, the total payoff of users will be depending on the specific allocation of purchase. To perform an optimal allocation that maximizes the users' payoff, the server needs full knolwdge about the privacy parameter $\lambda_i$ of each selling user $i$, which is assumed unavailable in our system model. Therefore, here we propose four privacy-knowledge-free strategies to handle such oversupply issue:
\begin{enumerate}
	\item \emph{Major sellers first}: The server will purchase data from users in a greedy manner and descending order upon their current supply $q_i^t$, until the demand is satisfied.
	\item \emph{Minor sellers first}: The server will purchase data from users in a greedy manner and ascending order upon their current supply $q_i^t$, until the demand is satisfied.
	\item \emph{Proportional}: The server will purchase data from all users, allocating its demand $\eta^t$ to every single selling user proportionally to their current supply $q_i^t$.
	\item \emph{Random order}: The server will purchase data from users in a greedy manner and random order.
\end{enumerate}

Remark that the oversupply will certainly trigger the satisfaction of server's optimal amount of data to keep $y\subscript{max}$, so it will only occur once, i.e., in the last round of the quotation.

% Figure here.

% Note that at each period $t$, by (\ref{equ:optimalselling}), we know that higher supply means lower privacy sensitivity (lower $\lambda_i$). Therefore, from the social welfare perspective, it is better to collect data form users with lower $\lambda_i$, as the cost of selling data for such users is lower too.

\subsection{Handling the Discontinuity in Cost}\label{subsec:discontinuity}
It must be recalled that the server's cost function $C(y)$ has a jump discontinuity at $y=d$, which is so far not considered in the above analyses.

When $y\subscript{max}<d$, due to this jump discountinuity, the actual server's cost of unlearning all user data of amount $d$ is $C(d)=\alpha A(d)$, which can be significantly lower than $C(y\subscript{max})$. Thus, after achieving $y^t=y\subscript{max}$, the server may still have an incentive to see if it worth buying all the remaining data from the users. Nevertheless, since the server does not have full knowledge about every user's privacy parameter, it cannot simply estimate the optimal price to offer. It is neither rational to simply continue purchasing more data at prices higher than the marginal cost of unlearning. To address this issue, we propose a post-quotation procedure, where after buying $y\subscript{max}$ amount of data, the server will still keep announcing new prices to the users like if the quotation continues, \emph{but not purchasing}. Instead, it observes if all users are willing to sell all remaining data at the new price. Only if so, the server will purchase all remaining data at the last quoted price, and terminates the post-quotation step. Otherwise, no purchase will be made even if the supply is non-zero, and the server simply updates the price. This procedure continues until the price $B^t$ is raised so high that the cost to buy all remaining data exceeds the unlearning cost saved therewith, i.e. when $B^t\cdot\left(d-y\subscript{max}\right)>C\left(y\subscript{max}\right)-C(d)$.

\subsection{Proposed Algorithm}
Summarizing the above discussions, we propose an iterative quotation-based price discovery protocol in Algorithm~\ref{alg:quotation}. Especially, for practical implementation in real use scanrios, we consider discrete user data sets with unit size of $\Delta d$, so that all trades of user data can only be executed by amount of integer times of $\Delta d$.

\SetKwProg{Pn}{Function}{:}{}%
\SetKwFunction{FMain}{Main}
\begin{algorithm}[!htpb]
	\caption{Iterative Price Discovery Protocol}
	% \caption{Buyer-Initiated English Auction}
	\label{alg:quotation}
	\scriptsize
	\DontPrintSemicolon
	\textbf{Input:} $a, A_1, A_2, A_3, T_0, \alpha, \beta, B^0, \Delta B, [d_1,d_2,\dots, d_I], \Delta d$,\;
	\textbf{Initialize:} $y^0\gets 0$, $t\gets 0$, $y\subscript{max}$ w.r.t. Eq.~\eqref{eq:ymax}, $\eta^0$ w.r.t. Eq.~\eqref{eq:demand}\;
	\While(\tcp*[f]{Quotation phase}){$\eta^t>0$}{
		Update $q_i^t$ for all $i\in\mathcal{I}$ w.r.t. Eq.~\eqref{equ:optimalselling}\;
		$q_i^t\gets \left\lfloor q_i^t/\Delta d\right\rfloor*\Delta d$\;
		\uIf{$\sum\limits_{i=1}^I q_i^t<\eta^t$}{
			$y_i^{t+1}\gets y_i^t+q_i^{t}$ for all $i\in\mathcal{I}$\;
		}
		\Else{
			Update $y_i^{t+1}$ for all $i\in\mathcal{I}$ upon the oversupply-handling strategy\;
		}
		$B^{t+1}\gets B^t+\Delta B$\;
		$\eta^{t+1}\gets \eta\left(y^{t+1},B^{t+1}\right)$\;
		$t\gets t+1$\;
	}
	\While(\tcp*[f]{Post-quotation phase}){$B^t\cdot\left(d-y^t\right)\leqslant C\left(y^t\right)-C(d)$}{
		Update $q_i^t$ for all $i\in\mathcal{I}$ w.r.t. Eq.~\eqref{equ:optimalselling}\;
		\If{$\sum\limits_{i=1}^I q_i^t=d-y$}{
			$y_i^{t+1}\gets d_i$ for all $i\in\mathcal{I}$\;
		}
		$B^{t+1}\gets B^t+\Delta B$\;
		$t\gets t+1$\;
	}
\end{algorithm}

\section{Numerical Evaluation}\label{sec:evaluation}
\subsection{Simulation Setup}
To verify our proposed approach, we conducted numerical simulation campaigns. The simulation setup is shown in Table~\ref{tab:setup}, following the specifications of \cite{CC2024price}.
\begin{table}[!htbp]
	\centering
	\caption{Simulation Setup}
	\label{tab:setup}
	\begin{tabular}{>{\cellcolor{white}}m{0.2cm} | m{1.8cm} m{2.7cm} m{2.4cm}}
		\toprule[2px]
%			\rowcolor{white}
		&\textbf{Parameter}&\textbf{Value}&\textbf{Remark}\\
		\midrule[1px]
		
		\rowcolor{gray!20}
		&	$I$	&	$10$	&	Number of users\\
		&	$d_i, \forall i\in\mathcal{I}$	&	6000 &	Data amount per user\\
		
		\rowcolor{gray!20}
		&	$\lambda_i, \forall i\in\mathcal{I}$	&	$\sim\mathcal{U}(0.5,29.5)$ &	User privacy\\		
		&	$\Delta d$	&	$1$	 &	Unit data amount\\

		\rowcolor{gray!20}
		&	$B^0$	&	$0.001$	 &	Initial price\\
		&	$\Delta B$	&	$0.001$	 &	Price step\\
		
		\rowcolor{gray!20}
		\multirow{-7}{*}{\rotatebox{90}{\textbf{System}}}&	$N$	&	$1000$ & Monte Carlo runs\\
		\midrule[1px]
		&	$[a, A_1,A_2,A_3]$	&	$[e, 0.1,3.33\times 10^{-5},0]$	&	Accuracy parameters\\		
		
		\rowcolor{gray!20}
		&	$T_0$	&	$2.85\times 10^{-4}$	&	Time factor\\
		\multirow{-3}{*}{\rotatebox{90}{\textbf{Cost}}} &	$[\alpha,\beta]$	&	[1500,1]	&	Weight factors\\
		
		\bottomrule[2px]
	\end{tabular}
\end{table}

\subsection{Oversupply Handling Strategies}
To compare the performance of the four oversupply-handling strategies proposed in Sec.~\ref{subsec:oversupply}, we carried out Monte Carlo tests. For each run, we measured the the server's payoff:
\begin{equation}
	\xi\subscript{s}=C^T-C(0)-\sum\limits_{t=0}^T B^t,
\end{equation}
where $T$ is the index of round when the procedure terminates, the users' total payoff
\begin{equation}
	\xi\subscript{u}=\sum\limits_{i=1}^I \left[\sum\limits_{t=1}^T B^{t-1} (y_i^t-y_i^{t-1})+U_i\left(y_i^T\right)-U_i(0)\right],
\end{equation}
and the social welfare
\begin{equation}
	\xi=\xi\subscript{s}+\xi\subscript{u}.
\end{equation}
The simulation results are shown in Table~\ref{tab:oversupply}. As we have expected, the selection of oversupply-handling strategy does not affect the server's payoff at all (difference within $0.2\%$), while the \emph{minor-seller-first} strategy outperforms the other three regarding the users' payoff and social welfare by a very slight superiority (within $2\%$). This is not suprising as the oversupply handling is triggered only in the last quotation round, which takes only a small portion of the total quotation procedure.
\begin{table}[!htbp]
	\centering
	\caption{Comparing the oversupply-handling strategies}
	\begin{tabular}{l|c|>{\columncolor{gray!20}}c|c|c}
			\toprule[2px]
			&\textbf{Major-first}&\textbf{Minor-first}&\textbf{Prop.}&\textbf{Rand.}\\
			\midrule[1px]
			\textbf{Server's payoff} & 684.8 & 683.6 & 684.1 & 684.9 \\
			\textbf{Users' payoff} & 1133.5 & \textbf{1158.2} & 1150.2 & 1134.0\\
			\textbf{Social welfare} & 1818.4 & \textbf{1841.8} & 1834.3 & 1818.9 \\
			\bottomrule[2px]
		\end{tabular}
	\label{tab:oversupply}
\end{table}

\subsection{Baseline Methods}
To comparatively benchmark the performance of our proposed quotation approach against state-of-the-art solutions, we also implemented three baseline methods:
\begin{enumerate}
	\item \emph{\Ac{dnr}}: This is a naive baseline where none of the users redeems any data, i.e., the server can keep all the user data for free, and therefore does not need to unlearn or purchase at all. However, all users undergo maximum privacy loss.
	\item \emph{\ac{gdpr}}: This is the rigid privacy solution, where no data trade option is provided, and the server must unconditionally unlearn all the user data upon request.
	\item \emph{\Ac{bsp}}: This is the solution proposed in \cite{CC2024price}, where the server solves an optimal universal price and quotes it to all users, each user then independently decides how much data to sell at that price.
\end{enumerate}
It is also worth to introduce the concept of \emph{informed ratio} following \cite{CC2024price}, which describes the ratio of users that are aware of their rights to ask for data redemption. The \emph{uninformed} users will be dealed with the \emph{\ac{dnr}} solution, no matter which solution is used for the \emph{informed} users.

\subsection{Benchmark Test}
\begin{figure}[!htbp]
	\centering
	\includegraphics[width=.9\linewidth]{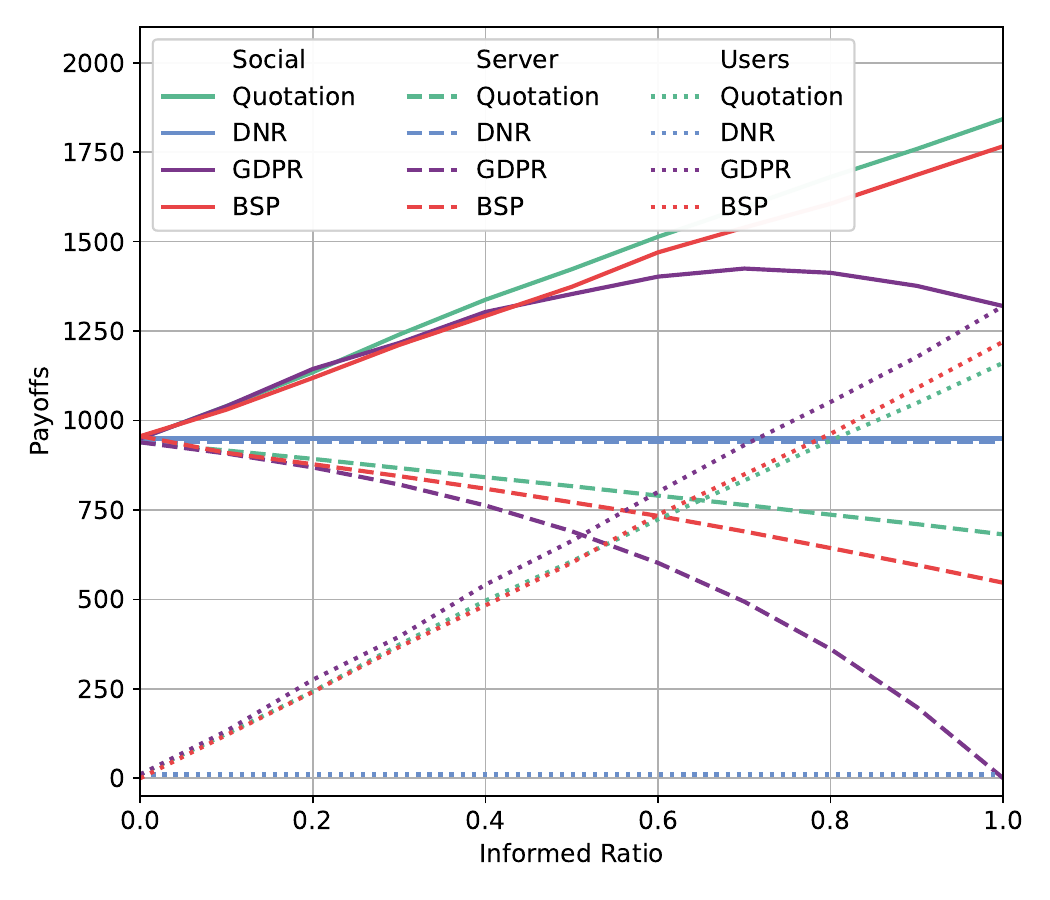}
	\caption{Performance of different data redemption solutions.}
	\label{fig:benchmark}
\end{figure}

We conducted a benchmark test to compare the performance of our proposed quotation approach with the aforementioned baseline methods. Specifically, the quotation approach was configured to handle oversupply with the \emph{minor-sellers-first} strategy.  We tested all methods at different informed ratios from $0\%$ to $100\%$ with $10\%$ step length, and measured the server's payoff, users' payoff, and social welfare. The results of Monte Carlo tests are shown in Fig.~\ref{fig:benchmark}. 

As we can see, our proposed quotation approach outperforms all the other methods in terms of social welfare. While the social welfare brought by the rigid \ac{gdpr} solution drops at high informed ratio due to the high unlearning cost, both our quotation approach and the \ac{bsp} solution exhibit an social welfare steadily increasing along with the informed ratio. Particularly, in comparison to \ac{bsp}, our method significantly improves the server's payoff at the cost of a less reduction in users' payoff, leading to its superiority in the overall social welfare. This shall be explained in two aspects: 
\begin{enumerate}
	\item The multi-round quotation mechanism allows the server to explore the users' non-linear privacy function over multiple intervals of data trading amount, which provides a closer approximation of the privacy utility loss by the payment. Compared to the \ac{bsp} solution that provides only one optimum \emph{average} price, the quotation mechanism is more cost-profitable for the buyer (server), but less profitable for the users.
	\item The convexity of of the server's cost function $C$ about the amount of data under keep $y$ guanrantees that the gain of the server's payoff is larger than the loss of the users' payoff.
\end{enumerate}

\section{Conclusions}\label{sec:conclusion}
In this paper, we have proposed an iterative quotation-based price discovery framework for data redemption in machine unlearning. We have analyzed the incentives of buying and selling data of the server and users, respectively, and derived the optimal selling strategy of users. Compared with state-of-the-art baselines, 
% our proposed solution does not require the server to possess any knowledge about the users' privacy parameter, and yet achieving a higher social welfare, which is demonstrated by numerical simulations.
our proposed quotation framework addresses a critical gap in deploying generative \ac{ai} within next-generation mobile networks: balancing user data rights with 
operational imperatives. By enabling \acp{mno} to dynamically compensate users for data retention, the mechanism sustains \ac{llm} accuracy--and thus network \ac{qos}--while respecting privacy preferences. Future work will explore: \begin{enumerate*}[label=(\arabic*)]
	\item taking into account users' benefit in their selling strategy regarding network performance improvement; and
	\item extending the model to scenarios with strategic information sharing among users regarding their data amount and privacy preferences.
\end{enumerate*}

\section*{Acknowledgment}
This work is supported by the Federal Ministry of Research, Technology and Space of Germany via the project Open6GHub+ (16KIS2406). B. Han (bin.han@rptu.de) is the corresponding author.

\bibliographystyle{IEEEtran}
\bibliography{references}

\end{document}

%% file: glossary.tex
\makeglossaries
\newacronym{genai}{GenAI}{generative artificial intelligence}
\newacronym{ai}{AI}{artificial intelligence}
\newacronym{bsp}{BSP}{bi-stage pricing}
\newacronym{ccpa}{CCPA}{\emph{California Consumer Privacy Act}}
\newacronym{dnr}{DNR}{do-not-redeem}
\newacronym{gdpr}{GDPR}{\emph{General Data Protection Regulation}}
\newacronym{llm}{LLM}{large language model}
\newacronym{ml}{ML}{machine learning}
\newacronym{mno}{MNO}{mobile network operator}
\newacronym{pipl}{PIPL}{\emph{Personal Information Protection Law}}
\newacronym{qos}{QoS}{quality of service}